\documentclass[10pt,twocolumn,letterpaper]{article}

\usepackage{iccv}
\usepackage{times}
\usepackage{epsfig}
\usepackage{graphicx}
\usepackage{amsmath}
\usepackage{amssymb}
\usepackage{subfigure}
\usepackage{multirow}
\usepackage{booktabs}
\usepackage{epstopdf}

\usepackage[breaklinks=true,bookmarks=false]{hyperref}

\iccvfinalcopy % *** Uncomment this line for the final submission

\def\iccvPaperID{****} % *** Enter the ICCV Paper ID here

\ificcvfinal\fi

\begin{document}

%%%%%%%%% TITLE
\title{SBSGAN: Suppression of Inter-Domain Background Shift for Person Re-Identification}

\author{Yan Huang$^\dag$$\quad$ Qiang Wu$^\dag$ $\quad$ JingSong Xu$^\dag$ $\quad$ Yi Zhong$^\S$\\
\dag School of Electrical and Data Engineering, University of Technology Sydney, Australia\\
\S School of Information and Electronics, Beijing Institute of Technology, China\\
{\tt\small \{yanhuang.uts, zhongyim2m\}@gmail.com,}
{\tt\small \{Qiang.Wu, JingSong.Xu\}@uts.edu.au}
}

\maketitle
% Remove page # from the first page of camera-ready.
\ificcvfinal\thispagestyle{empty}\fi

%%%%%%%%% ABSTRACT
\begin{abstract}
Cross-domain person re-identification (re-ID) is challenging due to the bias between training and testing domains. We observe that if backgrounds in the training and testing datasets are very different, it dramatically introduces difficulties to extract robust pedestrian features, and thus compromises the cross-domain person re-ID performance. In this paper, we formulate such problems as a background shift problem. A Suppression of Background Shift Generative Adversarial Network (SBSGAN) is proposed to generate images with suppressed backgrounds. Unlike simply removing backgrounds using binary masks, SBSGAN allows the generator to decide whether pixels should be preserved or suppressed to reduce segmentation errors caused by noisy foreground masks. Additionally, we take ID-related cues, such as vehicles and companions into consideration. With high-quality generated images, a Densely Associated 2-Stream (DA-2S) network is introduced with Inter Stream Densely Connection (ISDC) modules to strengthen the complementarity of the generated data and ID-related cues. The experiments show that the proposed method achieves competitive performance on three re-ID datasets, \ie, Market-1501, DukeMTMC-reID, and CUHK03, under the cross-domain person re-ID scenario.
\end{abstract}

%%%%%%%%% BODY TEXT
\section{Introduction}
\begin{figure}[t]
\begin{center}
\includegraphics[width=0.9\linewidth]{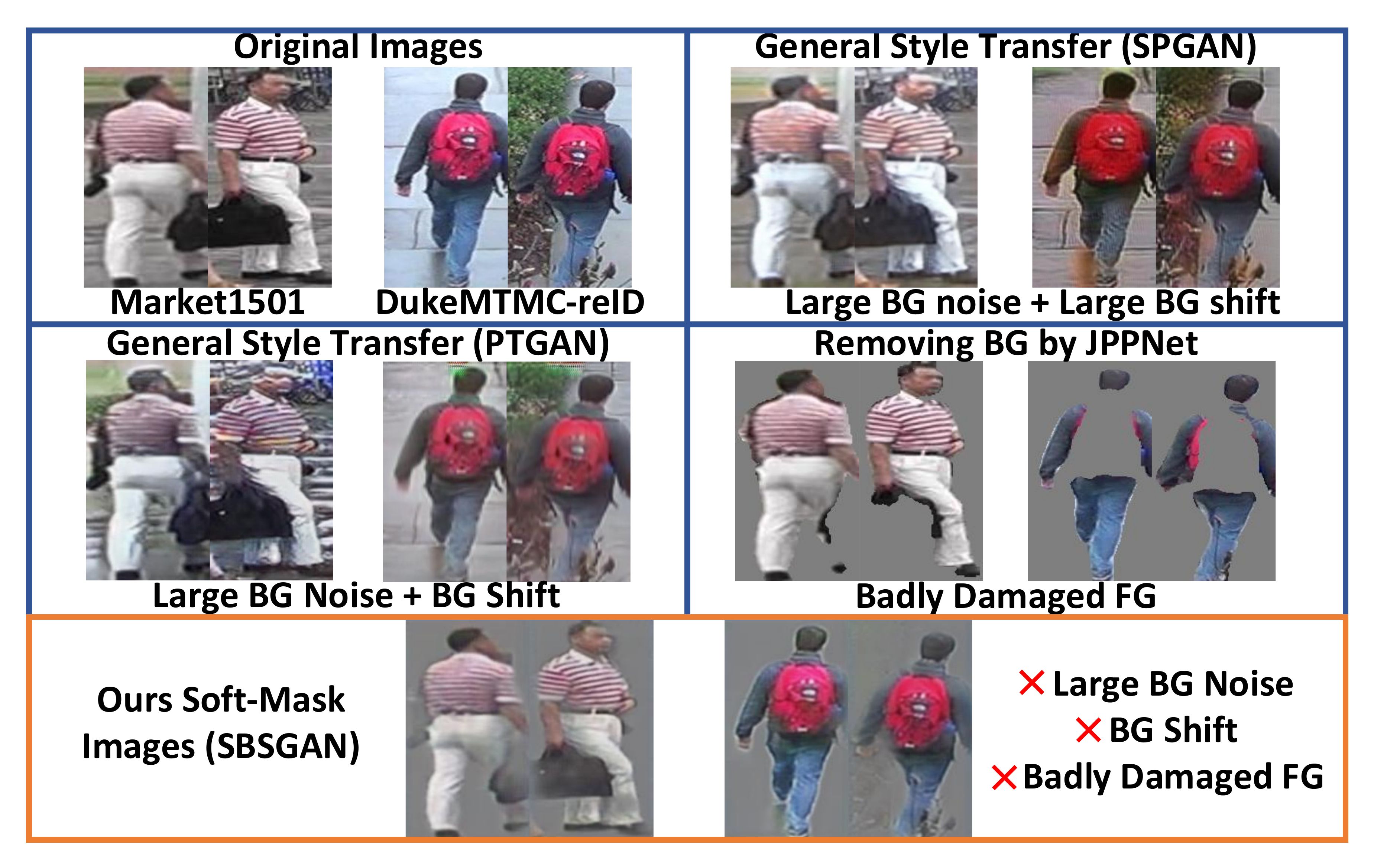}
\end{center}
   \caption{Comparison between different input images for cross-domain person re-ID. Images from Market-1501 and DukeMTMC-reID show distinct BG shift. Images generated by SPGAN~\cite{deng2018image} and PTGAN~\cite{wei2018person} do not suppress the BG noise, and have the BG shift problem. The hard-mask solution, \ie, JPPNet~\cite{liang2018look} damages the FG. Our SBSGAN takes all the impacts into consideration.}
\label{fig:intro}
\end{figure}

The task of person re-identification (re-ID) is to match the identities of a person under non-overlapped camera views~\cite{gong2014person,zheng2019joint,huang2017deepdiff,zheng2017unlabeled,li2014deepreid}. Most existing methods assume that the training and testing images are captured from the same scenario. However, this assumption is not guaranteed in many applications. For instance, person images captured from two different campuses have distinct illumination condition and background (BG) (\eg, Market-1501~\cite{zheng2015scalable} and DukeMTMC-reID~\cite{ristani2016performance,zheng2017unlabeled} datasets). In this situation, the bias between data distributions on two domains becomes large. Directly training a classifier from one dataset (\ie, source domain) often produces a degraded performance when testing is conducted on another dataset (\ie, target domain). Therefore, it is important to investigate solutions for such a cross-domain issue. For person re-ID, the domain adaption solutions have drawn attention in recent years~\cite{BakCL18,deng2018image,wang2018transferable,wei2018person,zhong2018generalizing}.

Recent cross-domain person re-ID methods usually adopt (or resort to) Generative Adversarial Network (GAN) to learn the domain variants~\cite{BakCL18,deng2018image,wei2018person,zhong2018generalizing}. These approaches can be categorized into two main types: 1) general inter-domain style transfer~\cite{BakCL18,deng2018image,wei2018person}; 2) inter-camera style transfer~\cite{zhong2018generalizing}. All of them may perform well on certain cases, \ie, domain style changes or camera style changes. However, they do not consider to remove or suppress BGs for reducing domain gaps. For instance, when a network is trained based on limited BG information presented in a source domain, such network may not well distinguish essential pedestrian features against noise caused by BG variations in a target domain. Unfortunately, BGs in the target domain is normally very different from the source domain. In this paper, we formulate this problem as a BG shift problem which may significantly degrade the overall performance of cross-domain person re-ID.

One possible solution to sort out BG shift is to directly remove BGs using foreground (FG) masks in a hard manner (\ie, applying the binary masks on original images)~\cite{farenzena2010person,huang2016person,song2018mask,tian2018eliminating}. However, it is observed that methods, such as JPPNet~\cite{liang2018look} and Mask-RCNN~\cite{matterport_maskrcnn_2017,he2017mask}, specifically being designed for removing BG may damage the FG information. By simply removing BGs, this hard manner solution does improve the performance of cross-domian person re-ID to a certain extent (see Table~\ref{tab:baseline}). At the same time, it can be seen that this is still an open problem. ``Is there a way to better suppress BG shift to improve cross-domain re-ID performance?'' This paper makes the first effort to generate images while BGs being suppressed moderately instead of completely removing the BGs in a hard manner.

To address the problem above, a Suppression of BG Shift Generative Adversarial Network (SBSGAN) is proposed. Compared with hard-mask solutions, images generated by the proposed SBSGAN can be regarded as FG images, with BG being suppressed moderately. The generated images by SBSGAN can be called as \textit{soft-mask images}. In addition, previous works~\cite{deng2018image,wei2018person} show that keeping the consistency of image style between domains can improve the performance of cross-domain person re-ID. Such an idea is also integrated into our SBSGAN to further reduce the domain gap. Fig.~\ref{fig:intro} shows images selected from two different person re-ID datasets. The BGs are quite different. A model trained on one dataset may easily be biased on another one due to the BG shift problem mentioned above. Images generated by recent cross-domain re-ID approaches, such as SPGAN~\cite{deng2018image} and PTGAN~\cite{wei2018person} still present some undesirable results. If we directly use FG masks obtained by JPPNet~\cite{liang2018look} to zero out BGs, the FG can be badly damaged by the noisy masks. On the contrary, every pixel in our generated images are preserved in a soft manner. Fig.~\ref{fig:intro} shows that our SBSGAN generates visually better images which can further reduce the domain gap caused by BG shift.

In order to enhance FG information and better integrate ID-related cues into the network, we propose a Densely Associated 2-Stream (DA-2S) network. This work is to argue that certain context information, \eg, companions and vehicles in BG may also provide ID-related cues. Both images with suppressed BGs (our generated images) and images with full BGs are respectively fed into the two individual streams of DA-2S. Unlike previous 2-stream methods (\eg,~\cite{ahmed2015improved,chen2018person,zheng2017discriminatively}), we propose Inter-Stream Densely Connection (ISDC) modules as new components used between the two streams of DA-2S. With ISDCs, more gradients produced by the final objective function can participate to strengthen the relationship between signals coming from two different streams in the back-propagation.

The contributions of this paper can be summarized in three-fold. 
1) BG shift is comprehensively investigated as an impact on cross-domain person re-ID. A SBSGAN is proposed to make the first effort by generating soft-mask images in order to reduce domain gaps. Compared with previous methods, BGs are mitigated rather than completely removed in our generated images.
2) A DA-2S CNN network with the proposed ISDC components is presented to facilitate complementary information between our generated data and more ID-related cues from the BG.
3) A comprehensive experiment is given to show the effectiveness of our soft-mask images in reducing domain gaps as well as the DA-2S model for cross-domain person re-ID.

%-------------------------------------------------------------------------
\section{Related Work}
Recently, followed by image-to-image translation approaches (\eg, CycleGAN~\cite{zhu2017unpaired} and StarGAN~\cite{choi2018stargan}), some researches focus on the inter-domain style transfer to reduce domain gaps for person re-ID. Deng \etal~\cite{deng2018image} proposed SPGAN to transfer general image style between domains. Wei \etal~\cite{wei2018person} introduced PTGAN to transfer the body pixel values and generate new BGs with the similar statistic distribution of the target domain. Unlike SPGAN, PTGAN explicitly considered the BG shift problem between domains. However, PTGAN overlooked the fact that BGs should be suppressed rather than retained, because the BG shift may degrade the cross-domain re-ID performance. In addition to the inter-domain style transfer, Zhong \etal~\cite{zhong2018generalizing} proposed to transfer the style of images between cameras to reduce the domain gap by using StarGAN~\cite{choi2018stargan}. A synthetic dataset was proposed to generalize illumination between different light conditions for cross-domain person re-ID in~\cite{BakCL18}. Cycle-consistency translation of GAN was employed to retain identities of the synthetic dataset. Unlike the these approaches, our SBSGAN concentrates on the BG shift problem by generating soft-mask images amongst different domains. We also take the style consistency into consideration to further reduce the domain gap.

To deal with the BG shift problem, one possible solution is to completely remove BGs using the binary body mask obtained by semantic segmentation or human parsing methods. Currently, methods such as Mask-RCNN~\cite{he2017mask} and JPPNet~\cite{liang2018look} can obtain body masks with the pre-trained model on large-scale datasets, \eg, MS COCO~\cite{lin2014microsoft} and LIP~\cite{liang2018look}. However, masks obtained by these methods often contained errors due to reasons such as low-resolution person images and highly dynamic person poses. Directly using the noisy masks may further jeopardize the cross-domain re-ID performance. Instead, we make the first effort to suppress the BG noise by generating soft-mask images. Previous work such as~\cite{kalayeh2018human} embeded the concept of `soft' to learn more informative features by using probability maps of different body parts on the feature level. Our SBSGAN focuses on the data level that tries to deal with the BG shift problem for cross-domain person re-ID.

\textbf{2-Stream Models} have been used in many computer vision tasks~\cite{ahmed2015improved,chen2018person,sun2015deeply,zheng2017discriminatively,zheng2017unlabeled,huang2018multi}. Generally, the learning objective of 2-stream models are categorized into two types. One verified inputs of the two individual streams belonging to the same or different classes, \eg,~\cite{ahmed2015improved,zheng2017discriminatively,zheng2017unlabeled,huang2018multi} in person re-ID and~\cite{sun2015deeply} in face recognition. The other type tried to enrich the representation by considering the complementarity between the inputs, \eg,~\cite{chen2018person} in person search. We follow the latter type and propose a DA-2S model. Unlike the above-mentioned 2-Stream models, ISDC is introduced between two individual streams of our DA-2S to strengthen the inter-stream relationship and explore a stronger complementarity between input images.

%-------------------------------------------------------------------------
\section{SBSGAN for Soft-Mask Image Generation}\label{sec:SBSGAN}
\begin{figure*}[t]
\begin{center}
\includegraphics[width=\linewidth]{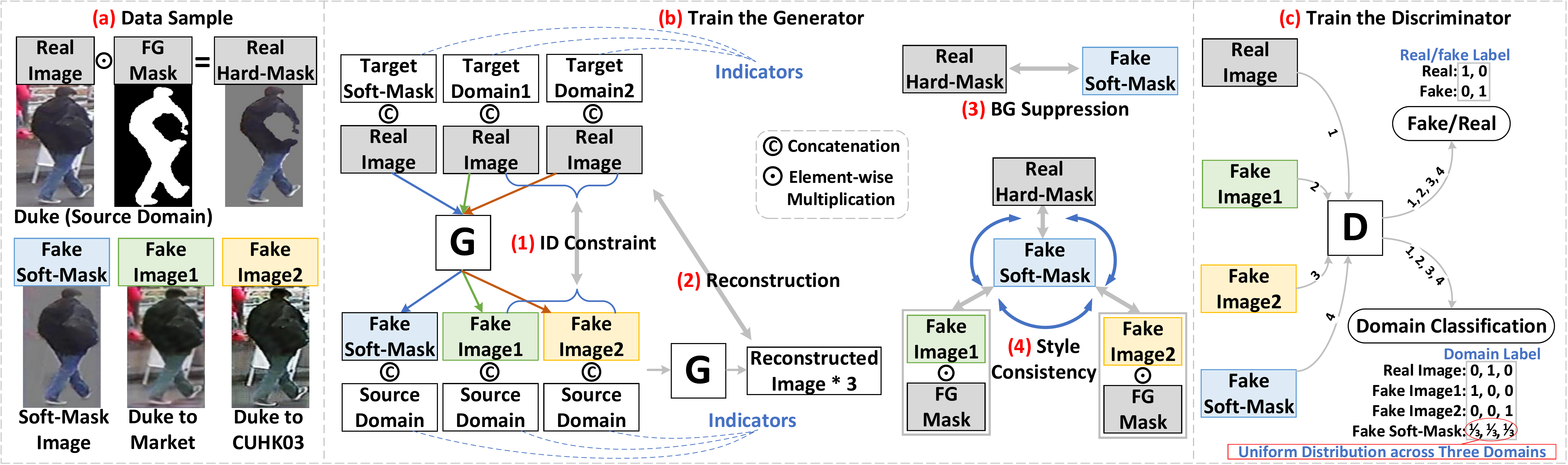}
\end{center}
   \caption{Overview of SBSGAN. Three domains are used as an example, including DukeMTMC-reID (source domain), Market-1501 (target domain1), and CUHK03 (target domain2). (a) shows an input image from DukeMTMC-reID. The FG mask is obtained by JPPNet. The generated soft-mask image and inter-domain style-transferred images are listed in the second row. (b) $G$ takes both input images and indicators as the inputs. All images (real/fake) participate in the different training process of $G$ to optimize different loss functions (1)-(4). (c) All the real and fake images are used to optimize the real/fake classification and domain classification losses in $D$.}
\label{fig:SBSGAN}
\end{figure*}

\subsection{Objective Functions in SBSGAN}
There are two tasks in the generator ($G$) of SBSGAN. The main task is to generate soft-mask images with suppressed BGs. The auxiliary task is to generate inter-domain style-transferred images (retain BG) to normalize the style of soft-mask images across all the training domains. Our discriminator ($D$) is used to distinguish the real and fake images, and classify these images to their corresponding domains. Fig.~\ref{fig:SBSGAN} shows the proposed SBSGAN.

Specifically, given an input image (\eg, $I_{\mathbb{D}_{s}}$) from source domain $\mathbb{D}_{s}$, $G$ can generate its corresponding soft-mask image $I_{\bar{\mathbb{D}}}$ by $G(I_{\mathbb{D}_{s}},\bar{\mathbb{D}})\rightarrow I_{\bar{\mathbb{D}}}$. $G$ takes both the input image (\eg, $I_{\mathbb{D}_{s}}$) and an indicator (\eg, $\bar{\mathbb{D}}$) as inputs. In addition, $G$ can also transfer the style of $I_{\mathbb{D}_{s}}$ to the $k$-th ($k\neq s$) target domain $\mathbb{D}_{k}$ via $G(I_{\mathbb{D}_{s}},\mathbb{D}_{k})\rightarrow I_{\mathbb{D}_{k}}$. The proposed SBSGAN supports multi-domain data as inputs. If there are $K$ domains in training, then, all $I_{\mathbb{D}_{k}} (k\in[1, K]\cap k\neq s$) and the input image $I_{\mathbb{D}_{s}}$ will be used to normalize the style of $I_{\bar{\mathbb{D}}}$, ensuring it is consistent across all the $K$ domains. Several loss functions are involved to train SBSGAN.

\textbf{(1) ID Constraint Loss.}
The ID constraint (IDC) loss was proposed to preserve the underlying image information (\eg, color) for data generation~\cite{taigman2016unsupervised}. We use IDC loss to preserve the color of person images for the auxiliary style-transferred image generation. The IDC loss is defined as follows:
\begin{equation}\label{equ:IDC}
\mathcal{L}_{idc} = \mathbb{E}_{I_{\mathbb{D}_{s}},\mathbb{D}_{k}}\left [\left \|G(I_{\mathbb{D}_{s}},\mathbb{D}_{k})-I_{\mathbb{D}_{s}}  \right \|_{1}\right ].
\end{equation}
We observe that without the IDC loss, $G$ may change the color of input images. Consequently, the color of generated soft-mask images are changed (see Fig.~\ref{fig:loss_abl}) when the auxiliary style-transferred images are directly applied to the soft-mask images for normalizing the style of them (see Eq.~\ref{equ:SC}).

\textbf{(2) Reconstruction Loss.}
We apply a reconstruction (REC) loss to ensure the content between an input image and its corresponding generated image remains unchanged. REC loss is a conventional objective function for the domain-to-domain image style transfer~\cite{choi2018stargan,deng2018image,wei2018person,zhu2017unpaired}. In our soft-mask image (or style-transferred image) generation, the image content of the FG (or FG+BG) should be kept with the input image. We only expect the domain-related parts being changed by the $G$. The REC loss is given as follows:
\begin{equation}\label{equ:rec}
\mathcal{L}_{rec} = \mathbb{E}_{I_{\mathbb{D}_{s}},\mathbb{D}_{k}\vee\bar{\mathbb{D}}}\left [\left \|G(G(I_{\mathbb{D}_{s}},\mathbb{D}_{k}\vee\bar{\mathbb{D}}),\mathbb{D}_{s})-I_{\mathbb{D}_{s}}  \right \|_{1}\right ],
\end{equation}
where $\vee$ is `or' operator.

\textbf{(3) BG Suppression Loss.}
We propose a BG Suppression (BGS) loss to suppress BG in data generation. The BGS loss also can preserve the FG color information of the generated soft-mask images. Therefore, part of functions between IDC loss and BGS loss are similar, but concentrate on generating different types of data. The BGS loss is formulated as follows:
\begin{equation}\label{equ:BGS}
\mathcal{L}_{bgs} = \mathbb{E}_{I_{\mathbb{D}_{s}},\bar{\mathbb{D}}}\left [\left \|I_{\mathbb{D}_{s}}\odot M(I_{\mathbb{D}_{s}}) - G(I_{\mathbb{D}_{s}},\bar{\mathbb{D}}) \right \|_{2}\right ].
\end{equation}
An auxiliary body mask $M(I_{\mathbb{D}_{s}})$ is used to suppress BG of the input image $I_{\mathbb{D}_{s}}$. $L_{2}$ distance is applied to minimize the loss. The JPPNet~\cite{liang2018look} is employed to extract $M(I_{\mathbb{D}_{s}})$. We find that masks obtained by JPPNet often contain segmentation errors. However, our SBSGAN is robust to the segmentation errors in the data generation process (see Fig.~\ref{fig:softhard}).

\textbf{(4) Style Consistency Loss.}
The Style Consistency (SC) Loss is proposed to encourage the style of soft-mask images (particular the part of FG) to be consistent across all the input domains, by which the domain gap of soft-mask images can be further reduced. The SC loss is given as follows: 
\begin{equation}\label{equ:SC}
\begin{aligned}
\mathcal{L}_{sc}& = \mathbb{E}_{I_{\mathbb{D}_{s}},\bar{\mathbb{D}},\mathbb{D}_{k}} [ \left \| G(I_{\mathbb{D}_{s}},\bar{\mathbb{D}})-I_{\mathbb{D}_{s}}\odot M(I_{\mathbb{D}_{s}}) \right \|_{1}+ \\
&\sum_{k=1,k\neq s}^{K}\left \| G(I_{\mathbb{D}_{s}},\bar{\mathbb{D}})-G(I_{\mathbb{D}_{s}},\mathbb{D}_{k})\odot M(I_{\mathbb{D}_{s}}) \right \|_{1}]. \\
\end{aligned}
\end{equation}
We first transfer the style of $I_{\mathbb{D}_{s}}$ to all the other $K-1$ domains. Then, $I_{\mathbb{D}_{s}}$ and all its corresponding style-transferred images are used to encourage the style of $G(I_{\mathbb{D}_{s}},\bar{\mathbb{D}})$ being consistent across all the $K$ domains.

Apart from the above-mentioned loss functions, we add the conventional adversarial loss ($\mathcal{L}_{adv}$)~\cite{goodfellow2014generative} of GAN to distinguish real and fake images in training. Also, $\mathcal{L}^{r}_{cls}$~\cite{choi2018stargan} is used to classify the source domains of real images for optimizing $D$, and $\mathcal{L}^{f}_{cls}$~\cite{choi2018stargan} is used to classify the target domains of fake images for optimizing $G$. Since the style of $I_{\bar{\mathbb{D}}}$ is normalized across all the $K$ domains, a uniform distribution (\ie, $\frac{1}{K}$) over the $K$ domains is assigned as the target domain of $I_{\bar{\mathbb{D}}}$.

Finally, the objective functions of $G$ and $D$ are respectively given as follows:
\begin{equation}\label{equ:softGAN_D}
\mathcal{L}_{D} = \mathcal{L}_{adv}+\mathcal{L}_{cls}^{r},
\end{equation}
\begin{equation}\label{equ:softGAN_G}
\begin{aligned}
\mathcal{L}_{G} = &\mathcal{L}_{adv}+\mathcal{L}_{cls}^{f}+\lambda_{rec}\mathcal{L}_{rec}+ \\
&\lambda_{idc}\mathcal{L}_{idc}+\lambda_{bgs}\mathcal{L}_{bgs}+\lambda_{sc}\mathcal{L}_{sc},
\end{aligned}
\end{equation}
where $\lambda$ is hyper-parameter to control the importance of different loss functions. We empirically set $\lambda_{rec}=10$ and $\lambda_{idc}=\lambda_{bgs}=\lambda_{sc}=5$ in all our experiments.

\subsection{Indicators in Data Generation}
The proposed SBSGAN supports multi-domain images as inputs. In experiments, images from three domains/datasets are used in training. When images are fed into $G$, an indicator is concatenated after each image on the dimension of channel to let $G$ knows which kind of image should be generated. A 3D tensor $\mathbb{D}$ is used as the indicator (see Fig.~\ref{fig:SBSGAN}). The height and width of $\mathbb{D}$ equal to the input image. There are $K$ channels in $\mathbb{D}$. For the auxiliary style-transferred image generation, $\mathbb{D}$ is denoted as $\mathbb{D}_{k}$; all values in the $k$-th channel of $\mathbb{D}_{k}$ are set to be one, and other values in the remaining $K-1$ channels are set to be zero. For the soft-mask image generation, $\mathbb{D}$ is denoted as $\bar{\mathbb{D}}$; all values of $\bar{\mathbb{D}}$ are set to be $\frac{1}{K}$.

\subsection{Network Architecture}
Adapted from~\cite{zhu2017unpaired}, given an input image, we use two down-sampling convolutional layers followed by six residual blocks~\cite{he2016deep} in $G$. Then, unlike~\cite{zhu2017unpaired}, two branches (without parameters sharing) are respectively used for generating soft-mask images and auxiliary style-transferred images followed by the output of the last residual block. Each branch contains two up-sampling transposed convolutional layers with the stride of 2. For $D$, we use the PatchGAN~\cite{isola2017image,zhu2017unpaired} structure.

%-------------------------------------------------------------------------
\section{Densely Associated 2-Stream Network}\label{sec:DA-2s}
\begin{figure}[t]
\begin{center}
\includegraphics[width=\linewidth]{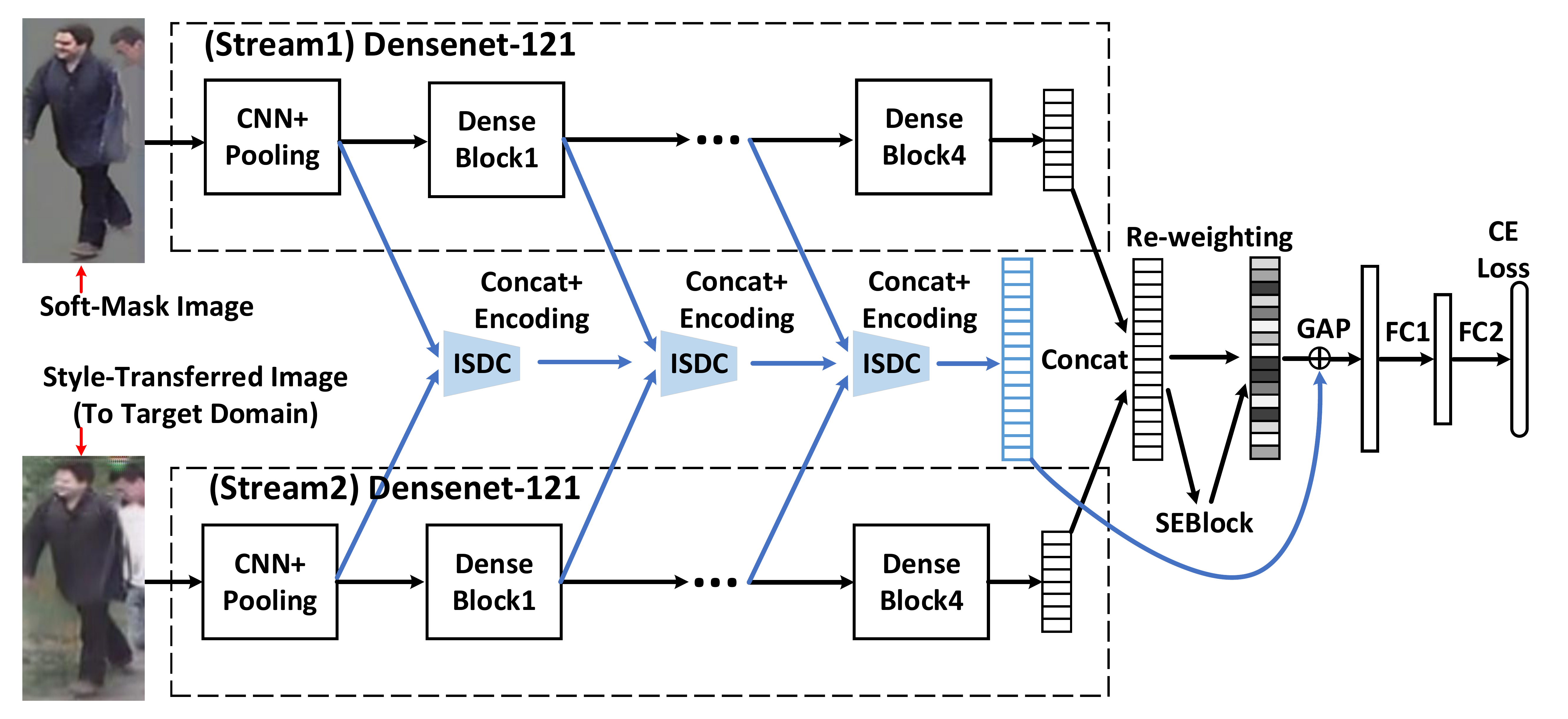}
\end{center}
   \caption{Overview of DA-2S. ISDC, GAP, FC, and CE respectively represent Inter-Stream Densely Connection, Global Average Pooling, Fully-Connected layer, and Cross-Entropy loss.}
\label{fig:2stream}
\end{figure}

The main contribution of this paper is to deal with the cross-domain person re-ID task from a brand new perspective, \ie, suppression of the inter-domain BG shift.
Moreover, to make use of helpful background cues, a DA-2S network is proposed. We argue that the context information, \eg, companions and vehicles in BG is also useful in cross-domain person re-ID. Therefore, our DA-2S network is used to enrich person representations by using both our soft-mask images and the image after general inter-domain style transfer. Fig.~\ref{fig:2stream} shows the DA-2S network. A pair of input images (a soft-mask image and its style-transferred image to the target domain) is fed into two ImageNet-trained Densenet-121~\cite{huang2017densely} networks (without parameters sharing). It can be observed that the companion in white clothes is regarded as BG being suppressed in the soft-mask image. To use the companion as an ID-related cue, a style-transferred images is fed into the second stream without suppressed BGs. To strengthen the complementarity of the two inputs, ISDC is proposed after the first pooling layer and every Dense Block. Specifically, the input information of each ISDC module is accumulated from both the outputs of the two streams as well as the previous ISDC module. Thus, the output of each ISDC module is defined as:
\begin{equation}\label{equ:ISDC}
O^{ISDC}_{n}=\delta (\mathcal{F}(y\cdot O^{ISDC}_{n-1}\oplus [O^{S1}_{n}, O^{S2}_{n}],\{\mathbf{W}_{n}\})),
\end{equation}
where $S1$ and $S2$ respectively represent the two streams, $O^{S1}_{n}$ and $O^{S2}_{n}$ are their respective output after the first pooling layer or each Dense Block, $n\in[1,4]$ represents the index of ISDC modules, $\mathcal{F}$ is a CNN encoder parameterized by $\mathbf{W}_{n}$, $\oplus$ is element-wise summation, $[\cdot]$ refers to concatenation along channel dimension, $y$ indicates whether this is the first (\ie, $n=1$) ISDC module between the two streams. If $n=1$, $y=0$, it refers to the first ISDC module. If $n\in [2,...,4]$, $y=1$, element-wise summation is used to transfer the knowledge from one previous ISDC module to the other. $\delta$ denotes ReLU~\cite{nair2010rectified}. Also, Batch Normalization (BN)~\cite{ioffe2015batch} is used before each ReLU activation function.

We re-weigh the final output of the two Densenet-121 backbone networks (after concatenation by channel) using SEBlock~\cite{hu2017squeeze} to emphasize informative features and suppress useless ones. The output of the last ISDC is directly connected to the re-weighted feature maps by an element-wise summation. Hence, gradients produced by objective function can be directly used to update parameters of layers connected to ISDC modules. Then, a global pooling is used followed by a fully-connected layer (FC1), BN, and ReLU. Another fully-connected layer (FC2) is used with $N$ neurons, where $N$ is the number of training identities. At last, a cross-entropy loss is adopted by casting the training process as an ID classification problem. Notably, we use DenseNet-121 as the backbone network because in each layer it takes all preceding feature maps as input to strengthen the gradients received by all preceding layers. The proposed ISDC module is also designed to strengthen the gradients produced by the inter-streams connections. We aim to verify whether the proposed ISDC module is still workable even with dense gradients being existed in the two individual streams (refer to Table~\ref{tab:ablation}).

%-------------------------------------------------------------------------
\section{Experiments}\label{sec:exp}
In this section, comprehensive evaluations (qualitative and quantitative) are carried out to verify the effectiveness of SBSGAN and DA-2S for cross-domain person re-ID. In the qualitative evaluation, we verify the effectiveness of soft-mask images generated by SBSGAN. In the quantitative evaluation, we evaluate our soft-mask images and DA-2S for cross-domain person re-ID. Our experiment is mainly conducted on Market-1501$\rightarrow$ DukeMTMC-reID (using Market-1501~\cite{zheng2015scalable} for training and DukeMTMC-reID~\cite{ristani2016performance,zheng2017unlabeled} for testing), since both datasets have fixed training/testing splits. In addition, other results are given on three widely used person re-ID datasets, including Market-1501, DukeMTMC-reID, and CUHK03~\cite{li2014deepreid}.

\subsection{Person Re-ID Datasets}\label{sec:dataset}
Table~\ref{tab:dataset} lists the training/testing settings of the three datasets. In the testing set, all query images are used to retrieve corresponding person images in the galley set. CUHK03 contains two image settings: one is annotated by hand-drawn bounding boxes, the other one is produced by a person detector. We only use and report the result of detected images which is more challenging. For all datasets, we use the single-query evaluation. The conventional rank-$n$ accuracy and mean Average Precision (mAP) are used as evaluation protocols~\cite{zheng2015scalable}.

\begin{table}[t]
\footnotesize
\centering
\caption{Person re-ID datasets for evaluations.}
\label{tab:dataset}
\begin{tabular}{c|c|c|c|c|c|c}
\toprule[1.3pt]
\multirow{2}{*}{Dataset} & \multicolumn{2}{c|}{Train} & \multicolumn{2}{c|}{Gallery (Test)} & \multicolumn{2}{c}{Query (Test)} \\  \cline{2-7}
 &\#ID &\#Img &\#ID &\#Img &\#ID &\#Img \\ \hline
Market~\cite{zheng2015scalable} & 751 & 12,936 & 750 & 19,732 & 750 & 3,368 \\
Duke~\cite{ristani2016performance,zheng2017unlabeled} & 702 & 16,522 & 702 & 17,661 & 702 & 2,228 \\
CUHK03~\cite{li2014deepreid} & 1,367 & 13,009 & 100 & 987 & 100 & 100 \\ \bottomrule[1.3pt]
\end{tabular}
\end{table}

\begin{figure*}[t]
\begin{center}
\includegraphics[width=\linewidth]{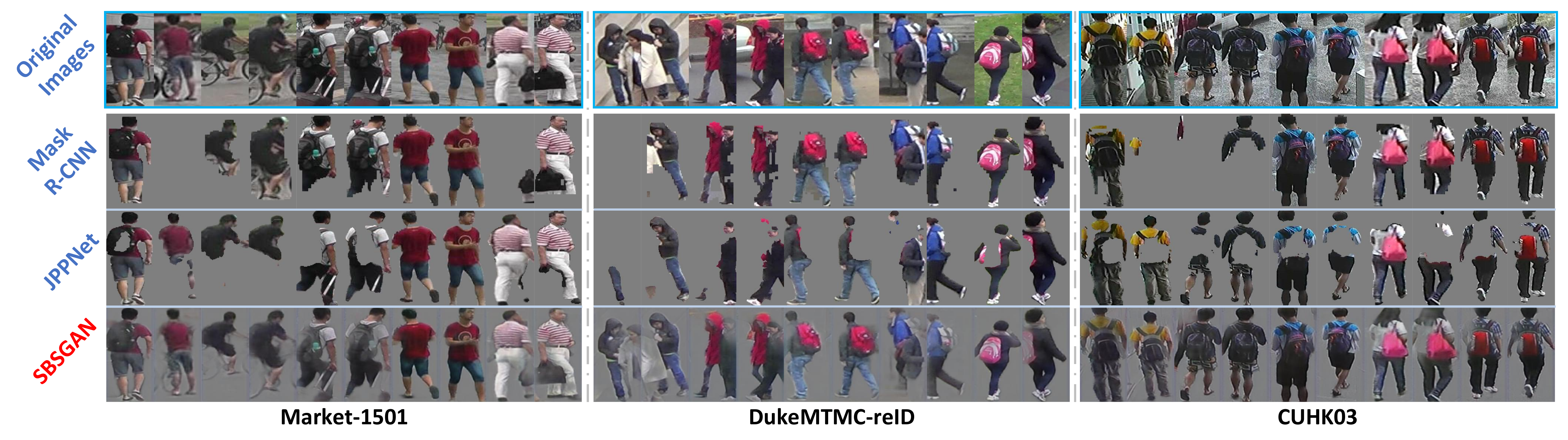}
\end{center}
  \caption{Comparison between hard-mask and soft-mask images. Images are selected from three different person re-ID datasets. The original images are listed in the first row. The second and the third rows respectively show hard-mask images by Mask-RCNN~\cite{matterport_maskrcnn_2017,he2017mask} and JPPNet~\cite{liang2018look}. The last row shows our soft-mask images generated by the proposed SBSGAN.}
\label{fig:softhard}
\end{figure*}

\subsection{Implementation Details}\label{sec:implementation}
\textbf{SBSGAN.}
All images of the three datasets ($K=3$) are used to train the proposed SBSGAN. Only weak domain labels are used. Input images and their corresponding body masks are resized to $256\times 128$. Adam~\cite{kingma2014adam} is used with $\beta_{1}=0.5$ and $\beta_{2}=0.999$. The batchsize is set to 16. To train $G$, $\frac{K+1}{16}$ images of each mini-batch are randomly selected for soft-mask images generation as well as the auxiliary style-transferred images generation. The remaining images in a mini-batch are used for the general style transfer to stabilize the performance of data generation in $G$. We initially set the learning rate to 0.0001 for both $G$ and $D$, and model stops training after 5 epochs. We perform one $G$ update after five $D$ updates as in~\cite{gulrajani2017improved}. In testing, an indicator (\ie, $\bar{\mathbb{D}}$) and an original image (\ie, $I_{\mathbb{D}_{s}}$) are concatenated for the soft-mask image generation. Notably, there is no need to use any FG or body mask in testing.

\textbf{DA-2S.}
Both soft-mask and style-transferred images (to the target domain) are used to train DA-2S (refer to Section~\ref{sec:DA-2s}). The soft-mask images are generated by the proposed SBSGAN. PTGAN~\cite{wei2018person} is used to get the general style-transferred images as the input to DA-2S. The batchsize is set to 50. Input images are resized to $256\times 128$ with random horizontal flipping. The SGD is used with momentum 0.9. The initial learning rate is set to 0.1 and decay to 0.01 after 40 epochs. We stop training after the 60-th epoch. A reduction rate of 16 is used for SEBlock as in~\cite{hu2017squeeze}. A dropout layer with the rate of 0.5 is inserted after FC1 (see Fig.~\ref{fig:2stream}) to reduce the risk of over-fitting. The FC1 has 512 neurons. According to the number of training identities, we set FC2 to have 751, 702, and 1,367 neurons when training is conducted on Market-1501, DukeMTMC-reID, and CUHK03 respectively. For each convolutional layer of ISDC, the kernel size=3, and padding=1. In addition, we use stride=2 for the first three ISDC modules and stride=1 for the last ISDC module. The number of channels is doubled by each ISDC. Finally, 2,048 channels are obtained after four ISDC modules. In testing, original images of the target domain and their corresponding soft-mask images are used as the inputs of DA-2S. We extract 2,048-dim CNN features for each testing image after the GAP layer. The Euclidean distance is used to compute the similarity between query and gallery images.

\subsection{Qualitative Evaluation}\label{sec:softresult}
\textbf{Soft-Mask Images Are Better Than Hard-Mask Images in Suppression of BG Shift.} In Fig.~\ref{fig:softhard}, we compare our soft-mask images with the hard-mask images. The hard-mask images are respectively obtained by JPPNet~\cite{liang2018look} and Mask-RCNN~\cite{matterport_maskrcnn_2017,he2017mask}. Both methods have shown compelling performance in person parsing or object instance segmentation. However, we find that the two methods cannot perform well in the segmentation of body from the BG on existing person re-ID datasets. It can be observed in Fig.~\ref{fig:softhard} that when people carry objects (\eg, bags), these objects are regarded as BGs and removed by noisy FG masks with segmentation errors. However, such features are significant to person re-ID, which should be retained rather than removed. In our soft-mask images, important cues such as bags and body parts can be well generated and retained. This is because we do not directly utilize the binary body mask on original images to remove the BGs. Although we use the mask obtained by JPPNet (the third row in Fig.~\ref{fig:softhard}) to suppress the BG (refer to Eq.~\ref{equ:BGS}) in data generation, our images visually show better results. This phenomenon also shows that the proposed SBSGAN is robust to the noisy masks in the data generation. 

\begin{figure}[t]
\begin{center}
\includegraphics[width=0.8\linewidth]{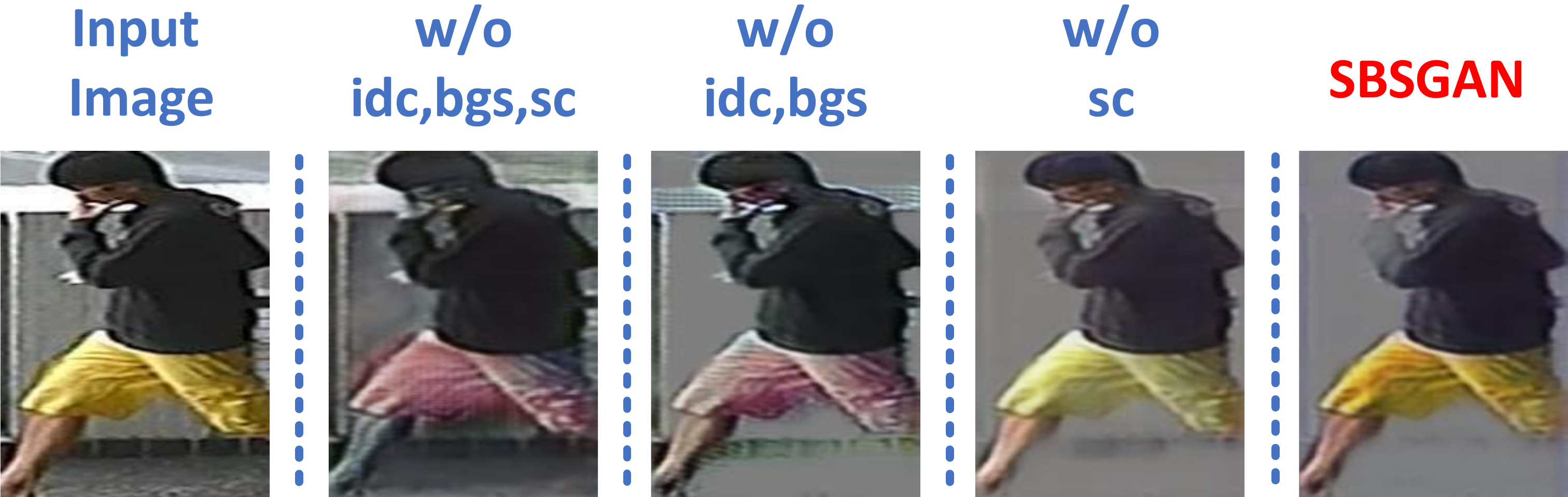}
\end{center}
   \caption{The effectiveness of different loss functions.}
\label{fig:loss_abl}
\end{figure}
\textbf{The Effectiveness of Loss Functions in SBSGAN.} The proposed SBSGAN jointly optimizes over several loss functions (see Eq.~\ref{equ:softGAN_D} and Eq.~\ref{equ:softGAN_G}). Fig.~\ref{fig:loss_abl} shows images generated by SBSGAN using different loss functions. We elaborate on the effectiveness of $\mathcal{L}_{idc}$, $\mathcal{L}_{bgs}$, and $\mathcal{L}_{sc}$. The others are conventional GAN-based loss functions, and their effectiveness is already verified by several previous works~\cite{arjovsky2017wasserstein,choi2018stargan,gulrajani2017improved,isola2017image,taigman2016unsupervised,zhu2017unpaired}. It can be observed in Fig.~\ref{fig:loss_abl} that when $\mathcal{L}_{idc}$ and $\mathcal{L}_{bgs}$ are removed, the color information of original images cannot be well preserved. In addition, the BG cannot be well suppressed. By only removing $\mathcal{L}_{sc}$, SBSGAN can generate soft-mask images which are close to our objective. The $\mathcal{L}_{sc}$ is proposed to encourage the style of generated soft-mask images being consistent (refer to Section~\ref{sec:SBSGAN}). Apart from the qualitative comparison in Fig.~\ref{fig:loss_abl}, a quantitative evaluation can be found in Table~\ref{tab:baseline} to further verify the effectiveness of $\mathcal{L}_{sc}$.

\begin{figure}[t]
\centering
\subfigure[Original images.]{
\includegraphics[width=3.95cm]{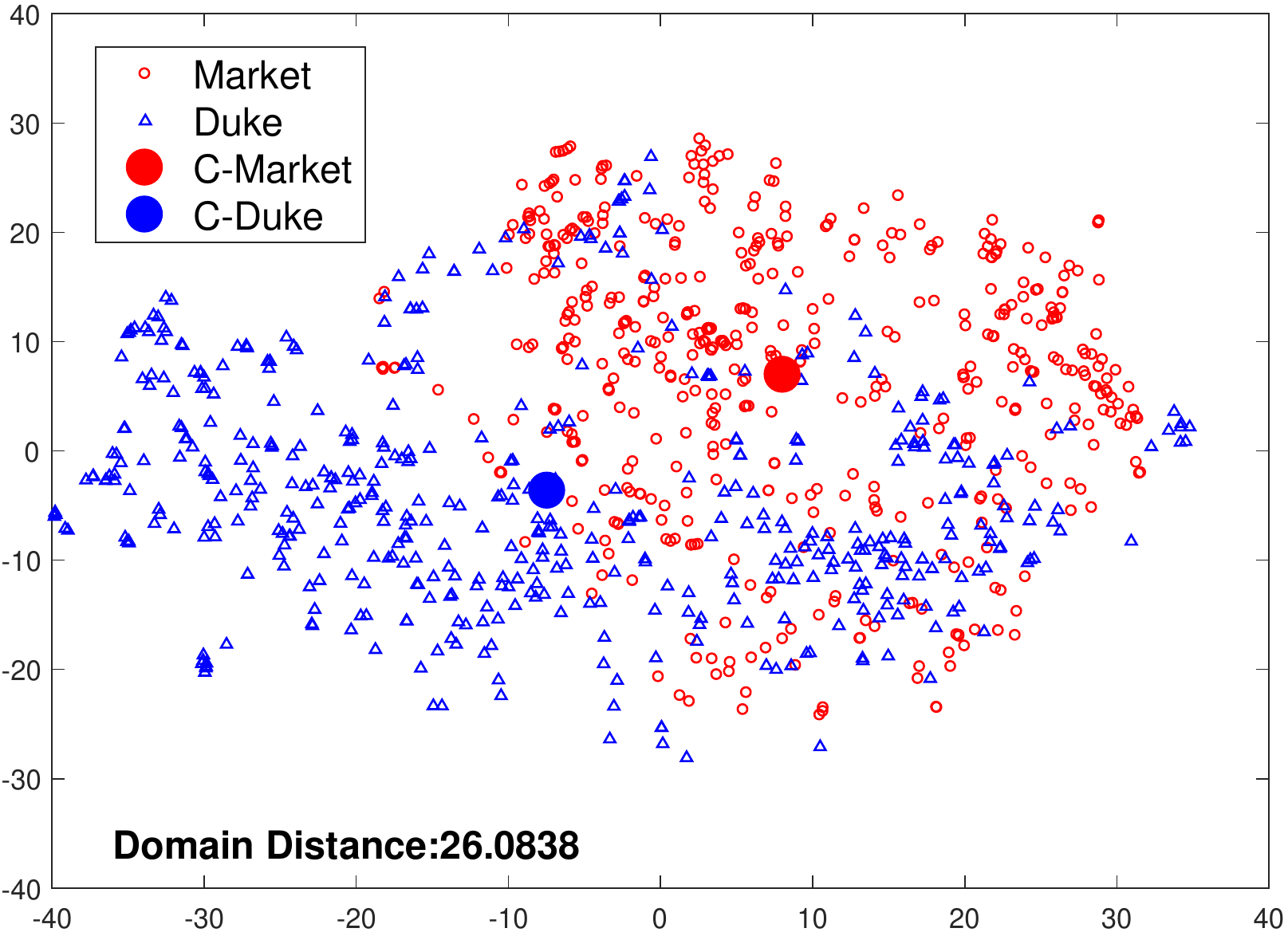}
}
\subfigure[StarGAN \cite{choi2018stargan}]{
\includegraphics[width=3.95cm]{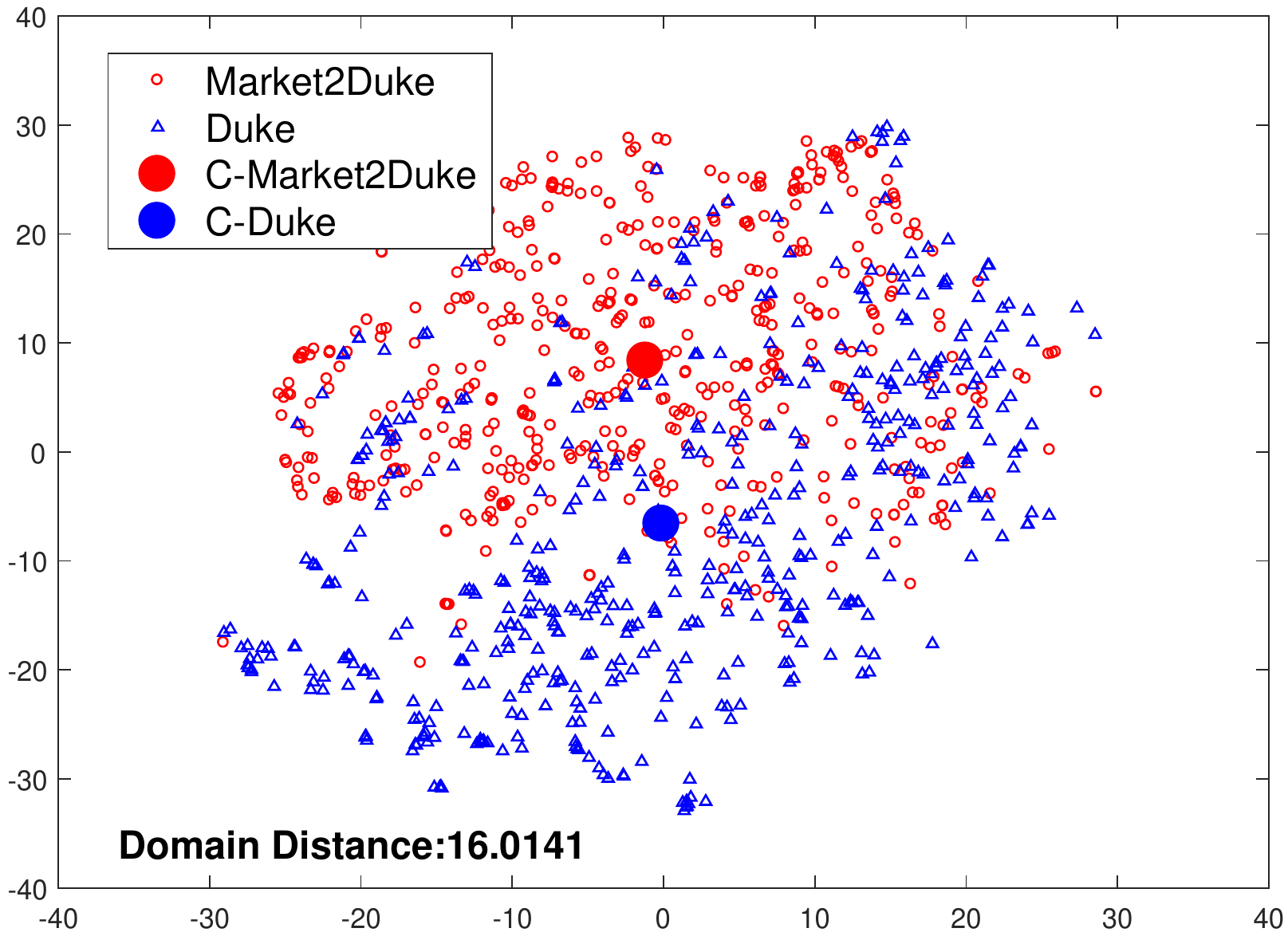}
}
\quad
\subfigure[SPGAN \cite{deng2018image}.]{
\includegraphics[width=3.95cm]{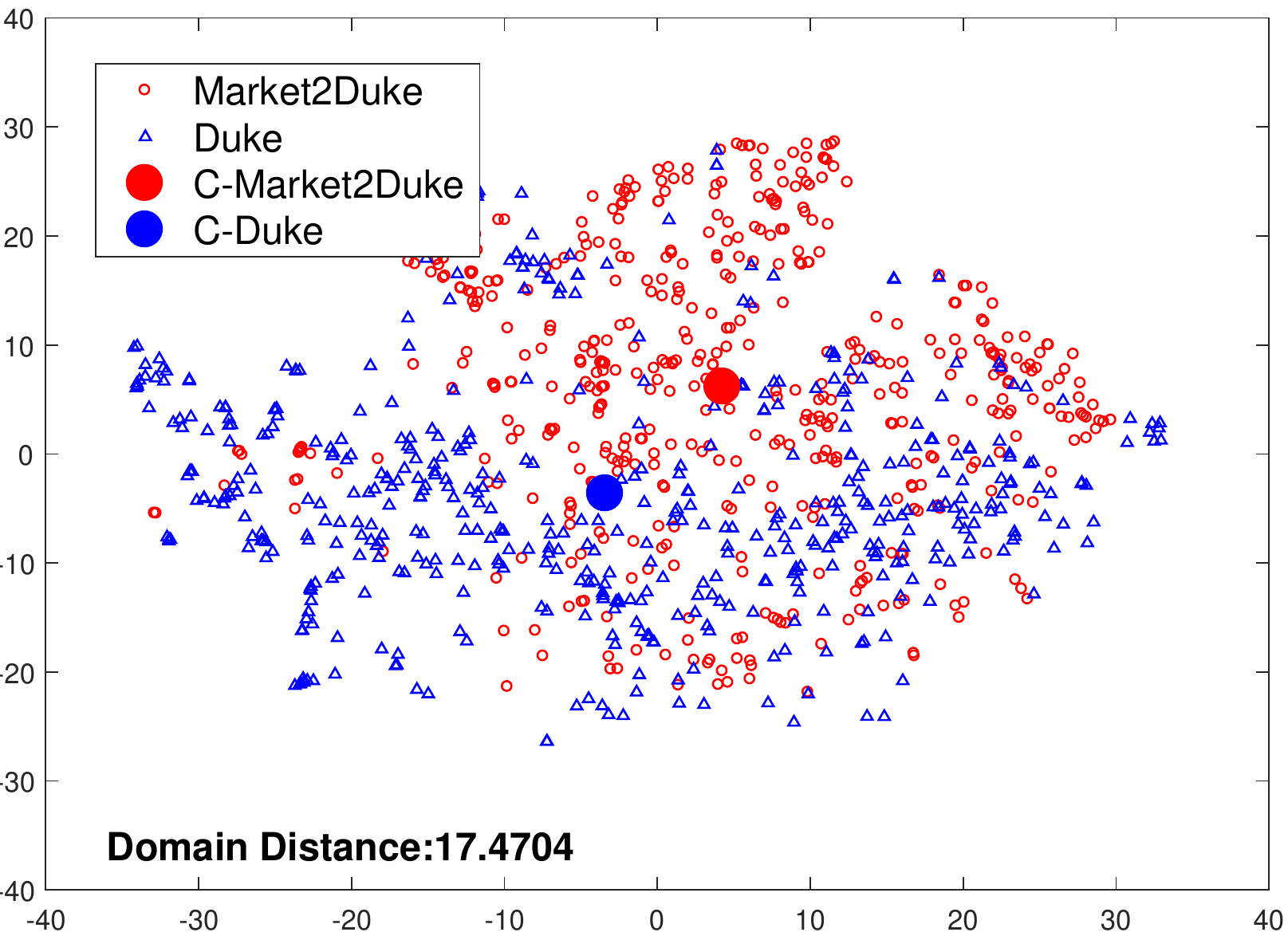}
}
\subfigure[PTGAN \cite{wei2018person}.]{
\includegraphics[width=3.95cm]{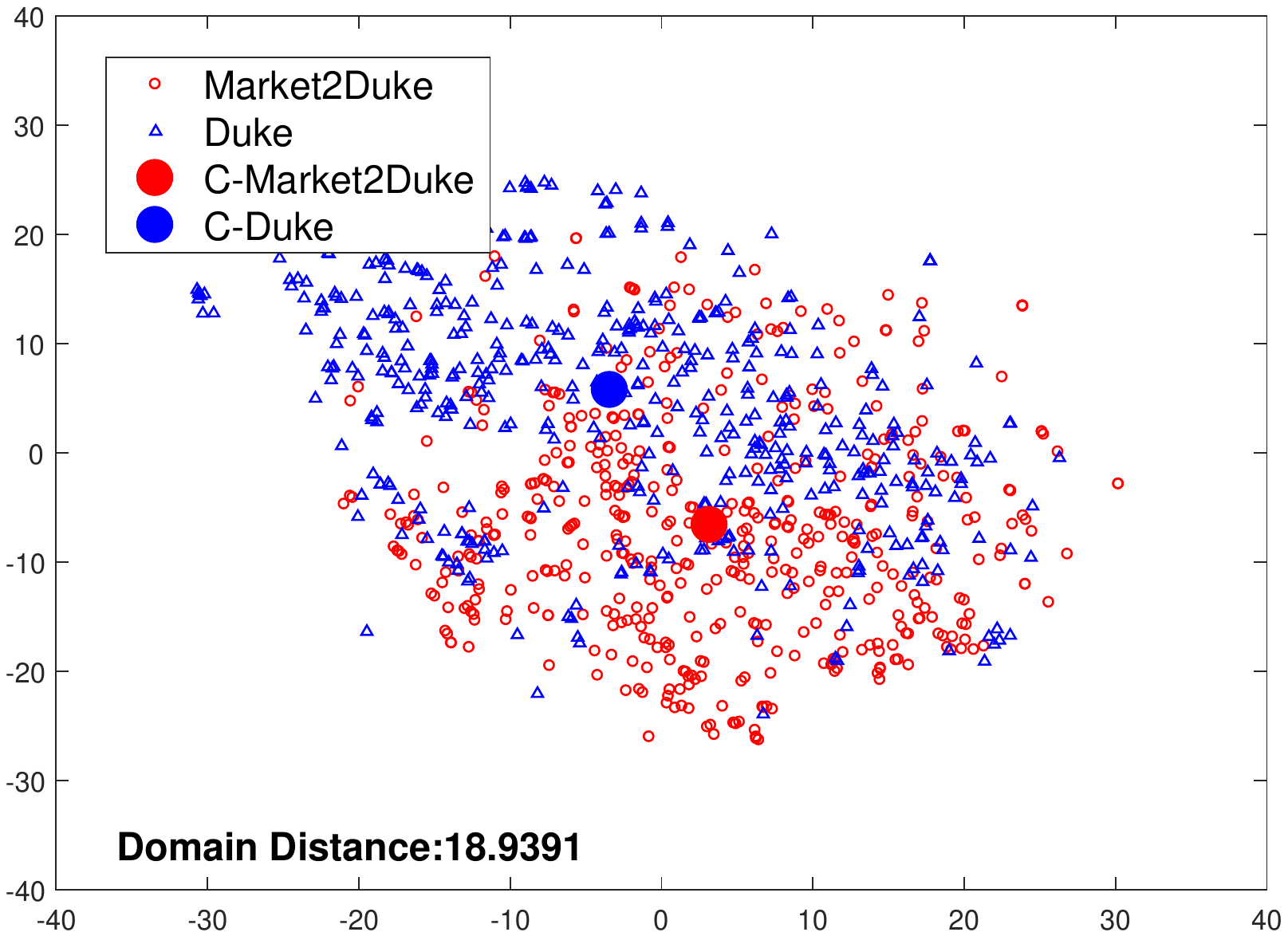}
}
\quad
\subfigure[Hard-mask images (JPPNet).]{
\includegraphics[width=3.95cm]{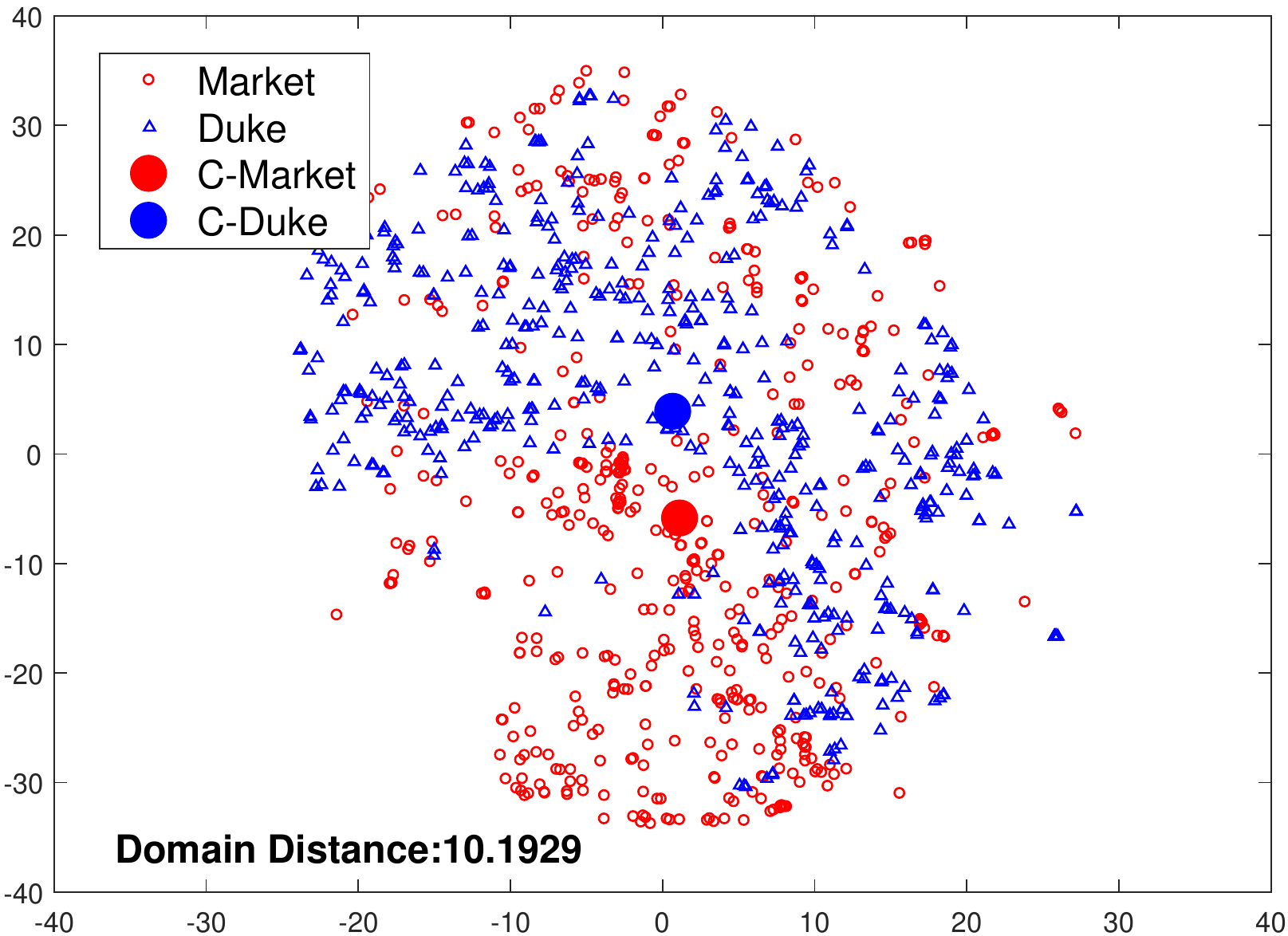}
}
\subfigure[\textcolor{red}{Soft-mask images (Ours).}]{
\includegraphics[width=3.95cm]{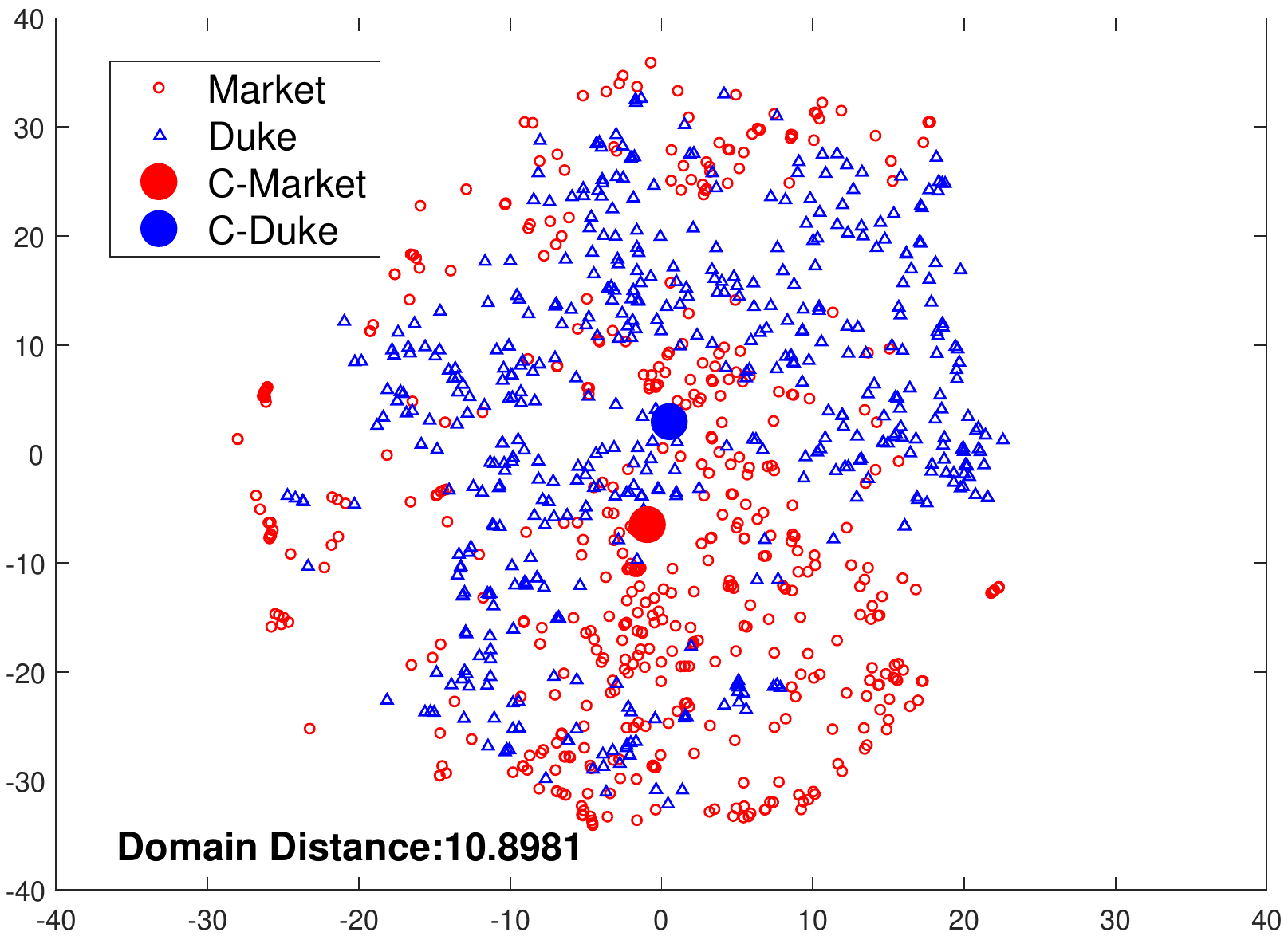}
}
\caption{Data visualization. 5000 images are randomly selected from Market-1501 and DukeMTMC-reID to learn data distributions via the Barnes-Hut t-SNE~\cite{Maaten14}, respectively. Another 200 images of each domain are used for visualization. The red circle and blue triangle respectively represent images belonging to Market-1501 and DukeMTMC-reID. The center points (\ie, `C-') are shown using their corresponding domain color. Domain distance (\ie, $L_{1}$ distance) is given between center points.}\label{fig:t-SNE}
\end{figure}
\textbf{Reducing the BG Shift Is Effective to Reduce Domain Gaps: Visualization of Data Distributions Between Two Domains.} We visualize the domain distance using different types of data, including the popular style-transferred images, hard-mask images, and our soft-mask images. Three recently published methods SPGAN~\cite{deng2018image}, PTGAN~\cite{wei2018person}, and StarGAN~\cite{choi2018stargan} are used to transfer the image style from Market-1501 to DukeMTMC-reID, respectively. Fig.~\ref{fig:t-SNE} shows the result. Compared with the general style-transferred results, the hard-mask and soft-mask images can reduce the domain gap by a large margin. This phenomenon verifies the effectiveness of reducing domain gaps by considering the BG shift problem. The domain distance of hard-mask images is on par with our soft-mask images (10.19 \vs 10.90). However, compared with hard-mask images, our soft-mask images show better performance in cross-domain person re-ID (\eg, rank-1: 43.3\% \vs 38.6\%, see Table~\ref{tab:baseline}). Naturally, it is unfair to directly compare the domain distance between soft-mask and hard-mask images. This is because many pixel values of hard-mask images are simply zeroed out, which makes approximately half the information of hard-mask images already being discarded in the comparison. Our soft-mask images suppress the BGs rather than simply removing them.
\begin{table}[t]
\footnotesize
\centering
\caption{Baseline performance of cross-domain person re-ID. Market-1501 is for training and DukeMTMC-reID is for testing.}\label{tab:baseline}
\begin{tabular}{c|c|c|c|c}
\toprule[1.3pt]
Training Data& mAP& R-1& R-5& R-10 \\ \hline
Original& 17.7 & 33.5 & 49.3 & 55.1  \\ \hline
\multicolumn{5}{c}{Hard-mask Images} \\ \hline
Mask-RCNN~\cite{matterport_maskrcnn_2017,he2017mask} & 20.6 & 37.5 & 53.4 & 59.1 \\
JPPNet~\cite{liang2018look} & 21.5 & 38.6 & 54.3 & 60.0 \\ \hline
\multicolumn{5}{c}{Style-transferred Images} \\ \hline
PTGAN~\cite{wei2018person} & 22.7 & 42.9 & 58.0 & 64.2 \\
SPGAN~\cite{deng2018image} & \textbf{\textcolor{blue}{22.8}} & 42.0 & 57.9 & 64.1 \\
StarGAN~\cite{choi2018stargan} & 21.6 & 39.8 & 53.4 & 59.9  \\ \hline
\multicolumn{5}{c}{Soft-mask Images (Ours)} \\ \hline
\textbf{Soft-mask w/o $\mathcal{L}_{sc}$} & 21.2 & 41.7 & 56.3 & 62.7 \\
\textbf{Soft-mask} & 22.3 & \textbf{\textcolor{blue}{43.3}} & \textbf{\textcolor{blue}{58.2}} & \textbf{\textcolor{blue}{64.4}} \\ 
\textbf{Soft-mask$_{2-Domains}$} & \textbf{\textcolor{red}{23.5}} & \textbf{\textcolor{red}{44.2}} & \textbf{\textcolor{red}{59.5}} & \textbf{\textcolor{red}{65.3}} \\ \bottomrule[1.3pt]
\end{tabular}
\end{table}

\begin{table}[t]
\footnotesize
\centering
\caption{Ablation study of DA-2S. Market-1501 is used for training and DukeMTMC-reID is used for testing. The baseline does not use SEBlocks and any ISDC modules. We also try to add SEBlocks to every ISDC module to re-weight the output of ISDC in the middle layers (denoted as ISDC-SE). The DA-2S$^\dag$ (DA-2S$^\ddag$) means only using the style-transferred images (soft-mask images) as the inputs of the 2-stream network.}
\label{tab:ablation}
\begin{tabular}{l|c|c}
\toprule[1.3pt]
Methods & mAP& R-1 \\ \hline
Basel. & 28.8 & 50.2 \\
Basel.+SEBlock & 28.9 & 50.5 \\
Basel.+SEBlock+ISDC-SE & \textbf{\textcolor{blue}{30.4}} & \textbf{\textcolor{blue}{51.5}} \\
\textbf{Basel.+SEBlock+ISDC (DA-2S)} & \textbf{\textcolor{red}{30.8}} & \textbf{\textcolor{red}{53.5}} \\ \hline
DA-2S$^\dag$ (2*Style-transfer) & 28.4 & 49.6 \\
DA-2S$^\ddag$ (2*Soft-mask) & 27.0 & 51.5 \\\bottomrule[1.3pt]
\end{tabular}
\end{table}

\begin{table*}[htbp]
\footnotesize
\centering
\caption{Comparison with state-of-the-art methods. M, C, and D respectively represent Market-1501, CUHK03, and DukeMTMC-reID. X$\rightarrow$Y means training is conducted on X and testing is conducted on Y.}\label{tab:comparison}
\begin{tabular}{l|c|c|c|c|c|c|c|c|c|c|c|c}
\toprule[1.3pt]
\multirow{2}{*}{Methods} & \multicolumn{2}{c|}{M$\rightarrow$ D} & \multicolumn{2}{c|}{M$\rightarrow$ C}&\multicolumn{2}{c|}{D$\rightarrow$ M}&\multicolumn{2}{c|}{D$\rightarrow$ C}&\multicolumn{2}{c|}{C$\rightarrow$ M}&\multicolumn{2}{c}{C$\rightarrow$ D} \\ \cline{2-13}
&mAP &R-1 &mAP &R-1 &mAP &R-1 &mAP &R-1 &mAP &R-1 &mAP &R-1  \\ \hline
UMDL~\cite{peng2016unsupervised} \textit{CVPR16} &7.3 &18.5 &- &- &12.4 &34.5 &- &- &- &- &- &- \\
CAMEL~\cite{yu2017cross} \textit{ICCV17} &- &- &- &- &26.3 &54.5  &- &- &- &- &- &- \\
PUL~\cite{fan2018unsupervised} \textit{TOMM18} &16.4 &30.0 &- &- &20.5 &45.5 &- &- &18.0 &41.9 &12.0 &23.0 \\
PTGAN~\cite{wei2018person} \textit{CVPR18} &- &27.4 &- &26.9 &- &38.6 &- &24.8 &- &31.5 &- &17.6 \\
SPGAN$_{LMP}$~\cite{deng2018image} \textit{CVPR18} & 26.4 &\textbf{\textcolor{blue}{46.9}} &- &- &26.9 &58.1 &-&- &-&- &-&-\\
TJ-AIDL~\cite{wang2018transferable} \textit{CVPR18}  &23.0  &44.3 &- &- &26.5 &58.2 &-&- &-&- &-&-\\
HHL~\cite{zhong2018generalizing} \textit{ECCV18}  &\textbf{\textcolor{blue}{27.2}}  &\textbf{\textcolor{blue}{46.9}} &- &- &\textbf{\textcolor{red}{31.4}} &\textbf{\textcolor{red}{62.2}} &-&-& \textbf{\textcolor{red}{29.8}} &\textbf{\textcolor{blue}{56.8}} &\textbf{\textcolor{blue}{23.4}} &\textbf{\textcolor{blue}{42.7}} \\ \hline
\textbf{DA-2S (Ours)} &\textbf{\textcolor{red}{30.8}} &\textbf{\textcolor{red}{53.5}} &\textbf{\textcolor{red}{32.5}} &\textbf{\textcolor{red}{42.2}} &\textbf{\textcolor{blue}{27.3}} &\textbf{\textcolor{blue}{58.5}} &\textbf{\textcolor{red}{27.3}} &\textbf{\textcolor{red}{33.7}} &\textbf{\textcolor{blue}{28.5}} &\textbf{\textcolor{red}{57.6}} &\textbf{\textcolor{red}{27.8}} &\textbf{\textcolor{red}{47.7}}\\ \bottomrule[1.3pt]
\end{tabular}
\end{table*}

\subsection{Quantitative Evaluation}\label{sec:evaluation}
\textbf{Soft-mask Images \vs Other Types of Images.}
The popular IDE model~\cite{deng2018image,zheng2017unlabeled} with ImageNet-trained DenseNet-121 as backbone network is adopted to compare our soft-mask images with the general style-transferred images and hard-mask images. Table~\ref{tab:baseline} lists the performance. By directly using the original images for cross-domain learning, the performance is inferior (mAP: 17.7\%, rank-1: 33.5\%). A clear performance improvement is achieved by simply removing BGs from both training and testing images using masks obtained by JPPNet and Mask-RCNN, respectively. However, the performance of our soft-mask images outperforms the hard-mask images by +4.7\% in rank-1 accuracy (43.3\% \vs 38.6\%). This is because hard-mask images often involve segmentation errors. General style-transferred results such as PTGAN and SPGAN achieve competitive performance. However, our soft-mask images obtain the best rank-1 accuracy (43.3\%), which shows their effectiveness by considering the BG shift problem in cross-domain person re-ID. In addition, without $\mathcal{L}_{sc}$, images generated by SBSGAN can satisfy the visual requirement (see Fig.~\ref{fig:loss_abl}), but the cross-domain re-ID performance is dropped by 1.1\% in mAP and 1.6\% in rank-1 accuracy. This is because we use $\mathcal{L}_{sc}$ to normalize the style of soft-mask images across multiple domains, by which the inter-domain gap can be further reduced. Since SPGAN and PTGAN only support images of two domains as inputs, we also train our SBSGAN in the same way instead of using images from three domains. Without interference from images of the third domain (\ie, CUHK03), we obtain performance gains by Soft-mask$_{2-Domains}$ (mAP: 23.5\%, rank-1: 44.2\%). However, we still use multiple domains as inputs in all the other experiments to generate soft-mask images. This is because we can train only one model instead of multiple models between any two domains.

\textbf{Ablation Study of DA-2S.}
An ablation study of our DA-2S network is given in Table~\ref{tab:ablation}. Without SEBlock and ISDC (\ie, baseline), we achieve 28.8\% in mAP and 50.2\% in rank-1 accuracy. By using SEBlock (similar to~\cite{chen2018person}), the performance is improved from 50.2\% to 50.5\% in rank-1 accuracy. To strengthen the inter-stream relationship, the baseline+SEBlock+ISDC produces the best performance (mAP: 30.8\%, rank-1: 53.5\%), demonstrating the effectiveness of the proposed ISDC modules. If we add SEBlock to every ISDC modules (ISDC-SE), the performance is dropped by 2\% in rank-1 accuracy. This is because additional SEBlocks produce more parameters which can potentially increase the risk of over-fitting. Moreover, we also change the inputs of our 2-stream DA-2S to style-transferred images or soft-mask images only (\ie, the network receives two style-transferred images or two soft-mask images). The results demonstrate that the combination of the two types of images is better than using them independently. 

\textbf{Comparison With State-of-the-Art Methods.}
We compare our method with several recently published state-of-the-art approaches, including three unsupervised methods, \ie, UMDL~\cite{peng2016unsupervised}, CAMEL~\cite{yu2017cross}, and PUL~\cite{fan2018unsupervised}, and four cross-domain re-ID approaches, \ie, PTGAN~\cite{wei2018person}, SPGAN+LMP~\cite{deng2018image}, TJ-AIDL~\cite{wang2018transferable}, and HHL~\cite{zhong2018generalizing}. For a fair comparison, all the selected cross-domain methods (including our method) use images from one domain/dataset for training the re-ID model and the other domain/dataset for testing; no extra training images or strong labels are used from the target domain.

Table~\ref{tab:comparison} lists the comparison results. It is clear to see that our DA-2S method achieves very competitive performance. For instance, on M$\rightarrow$D, our method outperforms the state-of-the-art method HHL by +3.6\% in mAP and +6.6\% in rank-1 accuracy; on C$\rightarrow$D, our performance is higher by +4.4\% in mAP and +5.0\% in rank-1 accuracy. Compared with our method, the HHL achieves the best performance on D$\rightarrow$M and competitive performance on C$\rightarrow$M. However, HHL uses extra camera labels in the target domain. Specifically, $N$ times images are generated according to the number of cameras to learn about the camera invariant features. This inherently limits its expansibility to the large camera networks (\eg, $N=100$), where the training data should be increased by $N$ (\eg, 100) times. Amongst all the methods, only PTGAN gives the performance on M$\rightarrow$C and D$\rightarrow$C. Under the same experimental setting, our DA-2S outperforms PTGAN by a large margin (+15.3\% and +8.9\% in rank-1 accuracy) when training is respectively conducted on Market-1501 and DukeMTMC-reID, and testing is conducted on CUHK03.

%-------------------------------------------------------------------------
\section{Conclusion}
In this paper, we verify that the BG shift problem can be considered to reduce domain gaps for cross-domain person re-ID. SBSGAN is proposed to generate soft-mask images with the BG being suppressed. Compared with hard-mask solutions, soft-mask images are able to suppress the BG in a moderate way. Compared with general inter-domain style-transferred approaches, soft-mask images can further reduce the domain gap by considering the BG shift problem. A DA-2S model is introduced along with the proposed ISDC module to make use of helpful background cues. Experiment results demonstrate the effectiveness of our method in both the qualitative and quantitative evaluations.

\section*{Acknowledgment}
This research is supported by an Australian Government Research Training Program Scholarship.

{\small
\bibliographystyle{ieee_fullname}
\bibliography{egbib}

\begin{thebibliography}{10}\itemsep=-1pt

\bibitem{matterport_maskrcnn_2017}
Waleed Abdulla.
\newblock Mask r-cnn for object detection and instance segmentation on keras
  and tensorflow.
\newblock \url{https://github.com/matterport/Mask_RCNN}, 2017.

\bibitem{ahmed2015improved}
Ejaz Ahmed, Michael Jones, and Tim~K Marks.
\newblock An improved deep learning architecture for person re-identification.
\newblock In {\em CVPR}, 2015.

\bibitem{arjovsky2017wasserstein}
Martin Arjovsky, Soumith Chintala, and L{\'e}on Bottou.
\newblock Wasserstein generative adversarial networks.
\newblock In {\em ICML}, 2017.

\bibitem{BakCL18}
Slawomir Bak, Peter Carr, and Jean{-}Fran{\c{c}}ois Lalonde.
\newblock Domain adaptation through synthesis for unsupervised person
  re-identification.
\newblock In {\em ECCV}, 2018.

\bibitem{chen2018person}
Di Chen, Shanshan Zhang, Wanli Ouyang, Jian Yang, and Ying Tai.
\newblock Person search via a mask-guided two-stream cnn model.
\newblock In {\em ECCV}, 2018.

\bibitem{choi2018stargan}
Yunjey Choi, Minje Choi, and Munyoung Kim.
\newblock Stargan: Unified generative adversarial networks for multi-domain
  image-to-image translation.
\newblock In {\em CVPR}, 2018.

\bibitem{deng2018image}
Weijian Deng, Liang Zheng, Guoliang Kang, Yi Yang, Qixiang Ye, and Jianbin
  Jiao.
\newblock Image-image domain adaptation with preserved self-similarity and
  domain-dissimilarity for person reidentification.
\newblock In {\em CVPR}, 2018.

\bibitem{fan2018unsupervised}
Hehe Fan, Liang Zheng, Chenggang Yan, and Yi Yang.
\newblock Unsupervised person re-identification: Clustering and fine-tuning.
\newblock {\em TOMM}, 2018.

\bibitem{farenzena2010person}
Michela Farenzena, Loris Bazzani, Alessandro Perina, Vittorio Murino, and Marco
  Cristani.
\newblock Person re-identification by symmetry-driven accumulation of local
  features.
\newblock In {\em CVPR}, 2010.

\bibitem{gong2014person}
Shaogang Gong, Marco Cristani, Shuicheng Yan, and Chen~Change Loy.
\newblock {\em Person re-identification}.
\newblock Springer, 2014.

\bibitem{goodfellow2014generative}
Ian Goodfellow, Jean Pouget-Abadie, Mehdi Mirza, Bing Xu, David Warde-Farley,
  Sherjil Ozair, Aaron Courville, and Yoshua Bengio.
\newblock Generative adversarial nets.
\newblock In {\em NeurIPS}, 2014.

\bibitem{gulrajani2017improved}
Ishaan Gulrajani, Faruk Ahmed, Martin Arjovsky, Vincent Dumoulin, and Aaron~C
  Courville.
\newblock Improved training of wasserstein gans.
\newblock In {\em NeurIPS}, 2017.

\bibitem{he2017mask}
Kaiming He, Georgia Gkioxari, Piotr Doll{\'a}r, and Ross Girshick.
\newblock Mask r-cnn.
\newblock In {\em ICCV}, 2017.

\bibitem{he2016deep}
Kaiming He, Xiangyu Zhang, Shaoqing Ren, and Jian Sun.
\newblock Deep residual learning for image recognition.
\newblock In {\em CVPR}, 2016.

\bibitem{hu2017squeeze}
Jie Hu, Li Shen, and Gang Sun.
\newblock Squeeze-and-excitation networks.
\newblock In {\em CVPR}, 2018.

\bibitem{huang2017densely}
Gao Huang, Zhuang Liu, Laurens Van Der~Maaten, and Kilian~Q Weinberger.
\newblock Densely connected convolutional networks.
\newblock In {\em CVPR}, 2017.

\bibitem{huang2016person}
Yan Huang, Hao Sheng, and Zhang Xiong.
\newblock Person re-identification based on hierarchical bipartite graph
  matching.
\newblock In {\em ICIP}, 2016.

\bibitem{huang2017deepdiff}
Yan Huang, Hao Sheng, Yanwei Zheng, and Zhang Xiong.
\newblock Deepdiff: Learning deep difference features on human body parts for
  person re-identification.
\newblock {\em Neurocomputing}, 2017.

\bibitem{huang2018multi}
Yan Huang, Jingsong Xu, Qiang Wu, Zhedong Zheng, Zhaoxiang Zhang, and Jian
  Zhang.
\newblock Multi-pseudo regularized label for generated data in person
  re-identification.
\newblock {\em TIP}, 2018.

\bibitem{ioffe2015batch}
Sergey Ioffe and Christian Szegedy.
\newblock Batch normalization: Accelerating deep network training by reducing
  internal covariate shift.
\newblock In {\em ICML}, 2015.

\bibitem{isola2017image}
Phillip Isola, Jun-Yan Zhu, Tinghui Zhou, and Alexei~A Efros.
\newblock Image-to-image translation with conditional adversarial networks.
\newblock In {\em CVPR}, 2017.

\bibitem{kalayeh2018human}
Mahdi~M Kalayeh, Emrah Basaran, Muhittin G{\"o}kmen, Mustafa~E Kamasak, and
  Mubarak Shah.
\newblock Human semantic parsing for person re-identification.
\newblock In {\em CVPR}, 2018.

\bibitem{kingma2014adam}
Diederik~P Kingma and Jimmy Ba.
\newblock Adam: A method for stochastic optimization.
\newblock In {\em ICLR}, 2015.

\bibitem{li2014deepreid}
Wei Li, Rui Zhao, Tong Xiao, and Xiaogang Wang.
\newblock Deepreid: Deep filter pairing neural network for person
  re-identification.
\newblock In {\em CVPR}, 2014.

\bibitem{liang2018look}
Xiaodan Liang, Ke Gong, Xiaohui Shen, and Liang Lin.
\newblock Look into person: Joint body parsing \& pose estimation network and a
  new benchmark.
\newblock {\em TPAMI}, 2018.

\bibitem{lin2014microsoft}
Tsung-Yi Lin, Michael Maire, Serge Belongie, James Hays, Pietro Perona, Deva
  Ramanan, Piotr Doll{\'a}r, and C~Lawrence Zitnick.
\newblock Microsoft coco: Common objects in context.
\newblock In {\em ECCV}, 2014.

\bibitem{nair2010rectified}
Vinod Nair and Geoffrey~E Hinton.
\newblock Rectified linear units improve restricted boltzmann machines.
\newblock In {\em ICML}, 2010.

\bibitem{peng2016unsupervised}
Peixi Peng, Tao Xiang, Yaowei Wang, Massimiliano Pontil, Shaogang Gong, Tiejun
  Huang, and Yonghong Tian.
\newblock Unsupervised cross-dataset transfer learning for person
  re-identification.
\newblock In {\em CVPR}, 2016.

\bibitem{ristani2016performance}
Ergys Ristani, Francesco Solera, Roger Zou, Rita Cucchiara, and Carlo Tomasi.
\newblock Performance measures and a data set for multi-target, multi-camera
  tracking.
\newblock In {\em ECCV}, 2016.

\bibitem{song2018mask}
Chunfeng Song, Yan Huang, Wanli Ouyang, and Liang Wang.
\newblock Mask-guided contrastive attention model for person re-identification.
\newblock In {\em CVPR}, 2018.

\bibitem{sun2015deeply}
Yi Sun, Xiaogang Wang, and Xiaoou Tang.
\newblock Deeply learned face representations are sparse, selective, and
  robust.
\newblock In {\em CVPR}, 2015.

\bibitem{taigman2016unsupervised}
Yaniv Taigman, Adam Polyak, and Lior Wolf.
\newblock Unsupervised cross-domain image generation.
\newblock In {\em ICLR}, 2016.

\bibitem{tian2018eliminating}
Maoqing Tian, Shuai Yi, Hongsheng Li, Shihua Li, Xuesen Zhang, Jianping Shi,
  Junjie Yan, and Xiaogang Wang.
\newblock Eliminating background-bias for robust person re-identification.
\newblock In {\em CVPR}, 2018.

\bibitem{Maaten14}
Laurens van~der Maaten.
\newblock Accelerating t-sne using tree-based algorithms.
\newblock {\em JMLR}, 2014.

\bibitem{wang2018transferable}
Jingya Wang, Xiatian Zhu, Shaogang Gong, and Wei Li.
\newblock Transferable joint attribute-identity deep learning for unsupervised
  person re-identification.
\newblock In {\em ICCV}, 2018.

\bibitem{wei2018person}
Longhui Wei, Shiliang Zhang, Wen Gao, and Qi Tian.
\newblock Person transfer gan to bridge domain gap for person
  re-identification.
\newblock In {\em CVPR}, 2018.

\bibitem{yu2017cross}
Hongxing Yu, Ancong Wu, and Weishi Zheng.
\newblock Cross-view asymmetric metric learning for unsupervised person
  re-identification.
\newblock In {\em ICCV}, 2017.

\bibitem{zheng2015scalable}
Liang Zheng, Liyue Shen, Lu Tian, Shengjin Wang, Jingdong Wang, and Qi Tian.
\newblock Scalable person re-identification: A benchmark.
\newblock In {\em ICCV}, 2015.

\bibitem{zheng2019joint}
Zhedong Zheng, Xiaodong Yang, Zhiding Yu, Liang Zheng, Yi Yang, and Jan Kautz.
\newblock Joint discriminative and generative learning for person
  re-identification.
\newblock In {\em CVPR}, 2019.

\bibitem{zheng2017discriminatively}
Zhedong Zheng, Liang Zheng, and Yi Yang.
\newblock A discriminatively learned cnn embedding for person reidentification.
\newblock {\em TOMM}, 2017.

\bibitem{zheng2017unlabeled}
Zhedong Zheng, Liang Zheng, and Yi Yang.
\newblock Unlabeled samples generated by gan improve the person
  re-identification baseline in vitro.
\newblock In {\em ICCV}, 2017.

\bibitem{zhong2018generalizing}
Zhun Zhong, Liang Zheng, Shaozi Li, and Yi Yang.
\newblock Generalizing a person retrieval model hetero-and homogeneously.
\newblock In {\em ECCV}, 2018.

\bibitem{zhu2017unpaired}
Junyan Zhu, Taesung Park, Phillip Isola, and Alexei~A Efros.
\newblock Unpaired image-to-image translation using cycle-consistent
  adversarial networks.
\newblock In {\em CVPR}, 2017.

\end{thebibliography}
}

\end{document}